# CEL: A Continual Learning Model for Disease Outbreak Prediction by Leveraging Domain Adaptation via Elastic Weight Consolidation

**Saba Aslam** [1,2], **Abdur Rasool** [1,2], **Hongyan Wu** [1,*] **Xiaoli Li** [1,3,4]

[1] Shenzhen Institute of Advanced Technology, Chinese Academy of Sciences, Shenzhen 518055, China
[2] Shenzhen college of Advanced Technology, University of Chinese Academy of Sciences, Beijing 100049, China
[3] Centre for Cognitive and Brain Sciences, University of Macau, Taipa, Macau SAR, China
[4] Faculty of Science and Technology, University of Macau, Taipa, Macau SAR, China
*Corresponding author: Hongyan Wu (hy.wu@siat.ac.cn)

***Abstract:*** Continual learning, the ability of a model to learn over time without forgetting previous knowledge and, therefore, be adaptive to new data, is paramount in dynamic fields such as disease outbreak prediction. Deep neural networks, i.e., LSTM, are prone to error due to catastrophic forgetting. This study introduces a novel CEL model for <u>c</u>ontinual <u>l</u>earning by leveraging domain adaptation via <u>E</u>lastic <u>W</u>eight <u>C</u>onsolidation (EWC). This model aims to mitigate the catastrophic forgetting phenomenon in a domain incremental setting. The Fisher Information Matrix (FIM) is constructed with EWC to develop a regularization term that penalizes changes to important parameters, namely, the important previous knowledge. CEL's performance is evaluated on three distinct diseases, Influenza, Mpox, and Measles, with different metrics. The high R-squared values during evaluation and reevaluation outperform the other state-of-the-art models in several contexts, indicating that CEL adapts to incremental data well. CEL's robustness and reliability are underscored by its minimal 65% forgetting rate and 18% higher memory stability compared to existing benchmark studies. This study highlights CEL's versatility in disease outbreak prediction, addressing evolving data with temporal patterns. It offers a valuable model for proactive disease control with accurate, timely predictions.

## 1. Introduction

Predicting disease outbreaks is crucial for timely healthcare response, resource allocation, and containment efforts. Early forecasts inform public awareness, guide research, and influence policy, benefiting public health and the economy. They also foster international collaboration in managing global health crises [1]. Deep learning algorithms have demonstrated their efficacy in predicting disease outbreaks [2]. Their hierarchical structure allows them to learn complex features from raw data, such as time-series data from disease surveillance or textual data from news reports [3]. However, these static deep learning models necessitate complete retraining when incorporating new data, leading to increased time, cost, and resource consumption, highlighting a deficiency in deep learning static models [4]. This is because these models are often designed for specific tasks or static datasets and face challenges when it comes to evolving data like disease outbreak prediction. In real-time, disease outbreak data comes sequentially and should be incremented in the model on a daily, weekly, monthly, or yearly basis, while existing prediction models work statically, which means they fail for the purpose they are generated for [4]. Unlike artificial neural networks, entities like humans and animals have the inherent ability to learn new things incrementally without forgetting previously learned skills. This emphasizes the need to improve artificial systems to mimic natural ones in this aspect [5].

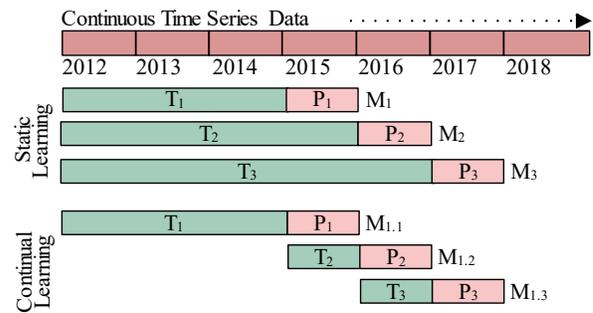

**Figure 1.** Static learning vs continual learning, T represents training data and P predictions.

To address this gap, the field of continual learning (CL) has emerged. A continual learning system is an adaptive model that learns from a continuous stream of information over time, without predefined task limits and without forgetting or interference when accommodating new information [6]. Its primary goal is to develop models that allow artificial systems to learn continuously over time without forgetting previous knowledge [4, 7]. The difference between static and continual learning is represented in Fig. 1, where in static learning, model ($M_1$) is trained on a fixed dataset ($T_1$) to give predictions ($P_1$), and every time new data ($T_2$, $T_3$) is incremented, a new model ($M_2$, $M_3$) is needed to be retrained, while CL model ($M_{1.1}$) can update its parameters in response to new data streams without the need for retraining from scratch.

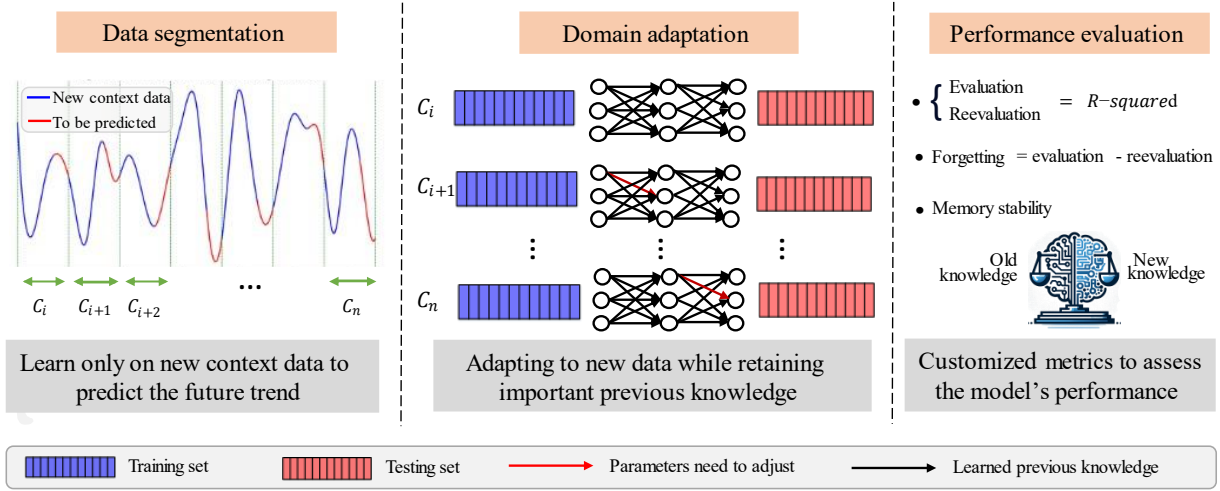

**Figure 2.** The architecture of the proposed model.

Interest in CL is rising due to its vast potential applications with deep neural networks. These applications range from medical diagnosis [8, 9] (where new diseases or symptoms might be discovered) to autonomous driving [10] (where the driving environment and rules might change) and real-time video surveillance [11] (which is a high-dimensional application). The mention of these applications emphasizes the practical importance and potential impact of advancements in CL. Across these diverse applications, the main challenge remains the effective mitigation of catastrophic forgetting (CF). This phenomenon, prevalent in neural networks, is characterized by the network's tendency to forget previously learned data when exposed to new data. More concretely, as these networks learn new tasks or adapt to fresh data distributions, the networks often lose the ability to execute tasks they were originally trained on. Such a situation poses a significant challenge, particularly when the expectation is for models to seamlessly assimilate new data without compromising their expertise on prior tasks [8].

Numerous studies [12-15] have been developed to solve the catastrophic forgetting problem. To devise robust solutions for combating CF, a fundamental understanding of the various types and scenarios within CL is important. It has three types [16] (i) task-incremental learning (Task-IL) which is the sequential and incremental adaptation of an algorithm to a series of discrete tasks [17, 18]. (ii) class-incremental learning (Class-IL), where an algorithm is faced with the evolving challenge of differentiating an expanding array of objects or classes [15, 19, 20], and (iii) domain-incremental learning (Domain-IL) that describes a situation where the problem structure remains constant, but the context or input distribution shifts, leading to changes in domains [21]. However, the majority of existing approaches are typically designed for Task-IL or Class-IL within the context of classification problems [11]. Whereas regression task is important in disease outbreak prediction as predicting the number of patients is essential for vaccine production and healthcare resource management, such as bed allocation. Moreover, there is a notable absence of research in Domain-IL [16] specifically targeting disease time series prediction within the framework of continual learning. This represents a gap in current research, indicating potential avenues for further investigations.

In this study, we developed a novel model called CEL for disease outbreak prediction designed to combat catastrophic forgetting in domain-incremental learning setting where the Fisher Information Matrix (FIM) in Elastic Weight Consolidation (EWC) is used to construct a regularization term that penalizes changes to important parameters for previous knowledge. The process starts with data segmentation for contextual learning, followed by domain adaptation where a neural network model incorporates with EWC and retains earlier knowledge while integrating new contexts. Finally, performance evaluation measures knowledge retention versus new learning. This architecture illustrated in Fig. 2, enables CEL to maintain historical insights while staying current with emerging data, optimizing its predictive accuracy for disease trends.

The contributions of this work are concluded as follows.

1. To the best of our knowledge, this research is pioneering in its attempt to develop a lightweight continual learning model for disease outbreak prediction by leveraging domain adaptation via EWC.
2. A novel data segmentation strategy is presented that partitions time-series data into discrete contexts, each encapsulating a distinct data distribution to facilitate domain-incremental learning.
3. The results consistently showed the CEL's effective performance in maintaining high R-squared values, forgetting, and memory stability metrics outperforming the other state-of-the-art models.
4. This is the first study that focuses on regression problems within continual learning and introduces the adapted forgetting measures.

The rest of the paper is organized as follows: Section 2 reviews existing approaches, Section 3 describes the methodology, Section 4 outlines the experiments, Section 5 discusses the results, and Section 6 concludes the paper.

## 2. Related Work

### 2.1 Deep Learning for Disease Outbreak Prediction

Deep neural networks have gained popularity among researchers in recent years for disease outbreak prediction. This is primarily due to their ability to extract meaningful features from input data. Venna [8] proposed a data-driven LSTM approach to explore the impact of environmental variables on real-time influenza prediction. Hybrid modeling approaches employ specialized optimization techniques during model tuning and have demonstrated effectiveness in influenza prediction [22, 23]. Kara [22] employed genetic algorithms to optimize LSTM model hyperparameters, enhancing multi-step influenza prediction accuracy. On the other hand, Yang [23] introduced a machine learning framework based on comprehensive learning particle swarm auto-regression chains, which extract patterns in various orders, resulting in improved performance compared to previous methodologies. Although transformers [24] are popular, their self-attention mechanism might not efficiently capture temporal dependencies in time series data. Additionally, they require large datasets to train without overfitting, which may not be available in case of disease outbreak data. However, these deep learning models suffer from catastrophic forgetting in disease outbreak prediction when facing new and incremental data under continual learning.

### 2.2 Continual Learning

Approaches for three scenarios of continual learning, Task-IL, Class-IL, and Domain-IL, fall under three categories: replay-based, regularization-based, and parameter isolation. Replay-based approaches maintain a limited exemplar memory to rehearse prior tasks while training new ones[12, 25-27]. Generative models can replace this memory to simulate past distributions [13, 28]. Techniques exist to minimize interference between tasks [14, 17]. Regularization-based approaches add terms to the training loss to restrict weight changes. Knowledge distillation is commonly used to maintain previous learning while adapting to new tasks [15, 25, 29]. Another method involves calculating the importance of each parameter and updating them accordingly [30, 31]. Parameter isolation mainly operates without model size constraints. It isolates important parameters from prior tasks, enabling new parameters for new tasks [32-34]. Zero-forgetting is achieved through learning masks or paths for each parameter or layer [35, 36]. However, most existing methods in CL are developed for Task-IL or Class-IL, focusing on classification challenges, while Domain-IL and regression tasks remain largely neglected.

## 3. Methodology

### 3.1 Problem Definition

A major characteristic of continual learning is its sequential learning process. We opted domain-incremental learning approach where the task (prediction) remains consistent, but data distribution undergoes shifts. At each time, only a small amount of the input data is available, called context $C$. Mathematically, domain-incremental learning for time series regression tasks can be expressed as follows:

Consider $N$ contexts $C = (C_1, \cdots, C_N)$ arriving sequentially. A context $C_i (i = 1,2,\cdots,N)$, consists of $k$ instances of labeled disease data $\mathcal{D}_i = \{(\mathbf{x}_{i,m}, y_{i,m})\}_{m=1}^{k}$, each time-series $\mathbf{x}_{i,m} \in \mathcal{X}_i$ with an associated target $y_{i,m} \in \mathcal{Y}_i$, and $m$ refers to the $m^{\text{th}}$ timepoint in the context $i$. The target space $\mathcal{Y}_i$ corresponds to real-valued numbers (or vectors). The goal in each context is to learn a solver model $M_i$ with trainable parameters $\boldsymbol{\theta}_i$ such that $M_i: \mathcal{X}_i \rightarrow \mathcal{Y}_i$, with an estimated target $\hat{y}_{i,m} = M_i(\mathbf{x}_{i,m}; \boldsymbol{\theta}_i)$ on condition of without accessing data from previous contexts $C_j (j = 1, \ldots, i-1)$.

### 3.2 Proposed Model

The proposed CEL model is designed to mitigate CF in Domain-IL, focusing on time-series data for multiple diseases. For a clear understanding of CEL, it is structured into three following distinct phases, as depicted in Fig. 3.

1. **Data segmentation Strategy:** We introduced a segmentation strategy by dividing the entire dataset into discrete $N$ contexts, each of which was split into both 80% training and 20% testing sets.
2. **Training CEL Model:** We constructed CEL model by training LSTM with EWC and FIM.

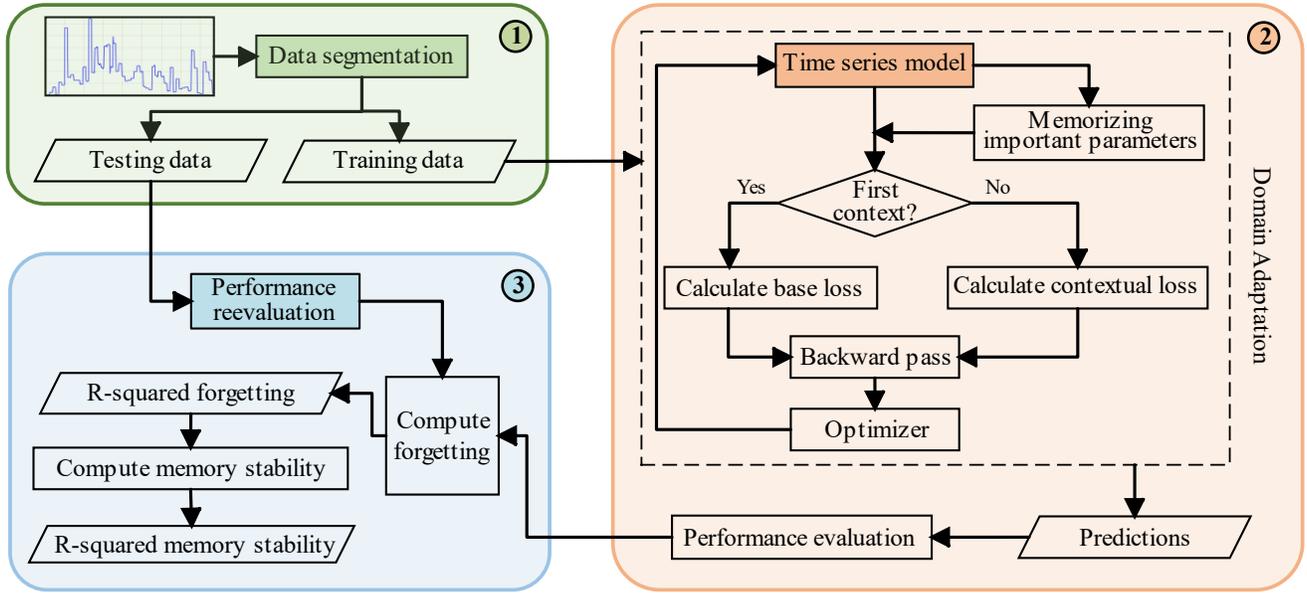

**Figure 3.** Proposed CEL model takes disease time series data and predicts outbreak, where ① presents data modeling including segmentation strategy, ② illustrates training model for prediction by mitigating catastrophic forgetting, and ③ shows schematic flow of performance metrics.

3. **CEL Performance Metrics:** We customized continual learning performance evaluation metrics [37] to assess the model's performance.

The following sub-sections explain the proposed model step by step.

### 3.2.1 Data Segmentation Strategy

To address the challenge posed by evolving data distributions, we adapted a segmentation strategy where we meticulously segmented our complete dataset into distinct $N$ contexts. Each context $\sum_{i=1}^{N} C_i$ (where N is the total number of contexts in the dataset) representative of unique data distribution was further divided into training (80%) and testing (20%) sets, as illustrated in Fig. 4. This segmentation strategy has two following objectives:

- By introducing diverse contexts and subjecting the model to both training and testing phases within each context, we simulate the model's exposure to evolving data distributions. This robustly gauges and augments its adaptability.
- This bifurcation ensures that as the model processes new data, its knowledge retention is periodically validated against the testing set, promoting consistent learning while preventing overfitting.

Time series disease data inherently shows variations, both temporally and geographically. Such a strategy is imperative for deriving comprehensive and accurate insights into the trajectory of the disease. This segmentation strategy revolves around selecting the optimal number of contexts $N$. For this, we partitioned the complete dataset into $N = (6, 7, ..., 10)$ contexts, executed a computational model on these segmented contexts, and the average R-squared values were computed for each configuration and finalized $N$ value such that.

1. More contexts allowed our model to capture the nuances in the data distribution, resulting in improved predictive performance.
2. Do not unnecessarily burden the model with an excessive number of contexts while still enabling effective learning.

### 3.2.2 CEL Training

In the second phase, we mitigate the catastrophic forgetting phenomenon by leveraging domain adaptation to new data.

*Domain Adaptation*

In a domain-incremental learning problem, the problem structure remains constant, but the context or input distribution shifts. Therefore, we need to leverage domain adaptation to new data while preventing the catastrophic forgetting of the knowledge from the older context. Drawing inspiration from the theory of plasticity of post-synaptic dendritic spines in the brain [5], Elastic Weight Consolidation (EWC) introduces a paradigm that determines the importance of a network parameter to previous contexts and provides a solution to the continual learning challenge by consolidating context-specific synaptic strengths (network parameters) [31]. It then penalizes any changes made to this parameter based on its importance while learning new contexts.

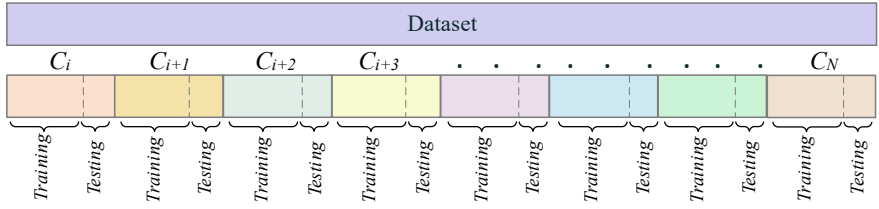

**Figure 4.** Data segmentation strategy, which segments the time series data into different contexts.

*Memorizing Important Parameters*

To explore how EWC addresses the challenges of approximating the posterior of $\theta$ in the context of deep learning models, the section paves the way for more effective learning across multiple contexts.

The training of deep neural networks is fundamentally an optimization problem. The objective is to find an optimal set of parameters, denoted by $\theta$, that minimizes the error in the training objective. Given the complexity of real-world contexts, there often exist multiple configurations of $\theta$ that can achieve near-optimal performance. In a continual learning scenario, where multiple contexts are learned sequentially, the challenge is to find a configuration of $\theta$ that performs well across all contexts. This requires the network to select parameters from the overlapping solution space of the all individual contexts [38] as represented in Fig. 5.

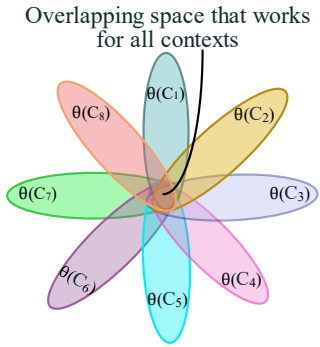

**Figure 5.** Example of 7 contexts, where EWC ensures that old contexts are remembered whilst training on new ones.

To regularize the appropriate parameters, EWC determines the importance of each parameter to that context once a context $C_i$ is learned. This importance is quantified using the Fisher Information Matrix (FIM) [39]. The FIM captures the sensitivity of the learned probabilistic model to changes in the parameters. The FIM ($F_i$) is typically calculated as the second derivative of the log-likelihood function with respect to the model parameters. In mathematical notation, it can be expressed as [38]:

$$F_i(\theta) = \mathbb{E}\left[\left(\frac{\partial}{\partial \theta} log\ p(x\mid \theta)\right)^2\right] \quad (1)$$

where $F_i(\theta)$ is the FIM regarding the model parameters $\theta$, $\mathbb{E}$ presents the expectation, which is usually calculated over the entire dataset, and $\left(\frac{\partial}{\partial \theta} log\ p(x\mid \theta)\right)^2$ is the second derivative of the log-likelihood function with respect to the model parameters $\theta$.

*Contextual Loss Formulation*

After learning a context, the network parameters are fine-tuned for subsequent contexts. However, changes to parameters deemed important for previous contexts are penalized. The penalty is proportional to the squared difference between the new and old values of the parameter, weighted by its importance. The new objective function, which includes this penalty, is [31]:

$$L(\theta) = L_{new}^{original}(\theta) + \sum_i \lambda/2 \left[F_i\left(\theta_i - \theta_{i,old}^*\right)^2\right] \quad (2)$$

where $L_{new}^{original}(\theta)$ is the original loss function for a new context, $\lambda$ is a hyperparameter that determines the strength of the regularization. $F_i$ is the Fisher Information for parameter $\theta_i$, which quantifies its importance to the old context. $\theta_{i,old}^*$ is the value of the parameter $\theta_i$ after training on old context. The term $\sum_i \lambda/2 \left[F_i\left(\theta_i - \theta_{i,old}^*\right)^2\right]$ acts as a regularization term. It penalizes changes to parameters that were important for the old context. The larger the Fisher Information $F_i$ for a parameter, the more it is penalized for deviating from its value after training in the old context.

In essence, while training on context, the network is not only trying to minimize the error for it but also ensuring that it doesn't forget the knowledge from the old context. This is achieved by adding a penalty to the loss function, which discourages significant changes to important parameters from the old context. This loss function is central to the EWC method and ensures that the network can learn new contexts without forgetting the old ones.

*Time Series Prediction*

We chose Long Short-Term Memory (LSTM) as a deep learning model for disease outbreak prediction because LSTM models are specifically designed to handle sequential data, making them particularly suitable for lightweight time series continual learning.

In LSTM cells, a crucial component termed the cell state $c_{t-1}$ acts as the long-term memory of the unit. Another vital aspect of the LSTM unit is the hidden state $h_{t-1}$, which symbolizes the short-term memory. In the sigmoid activation function, the output effectively dictates the proportion of long-term memory that is retained. It fundamentally influences the retention capacity of long-term memory and is aptly termed the forget gate.

$$f_l = \sigma\left(W_{l,h} * h_{t-1} + W_{l,x} * x_t + b_l\right) \quad (3)$$

where $f_l$ is the percentage of long-term to remember by forget gate.

Synthesis of potential long-term memory involves assimilating short-term memory and the current input. Concurrently, the network must determine the quantum of this potential memory to be incorporated into the long-term memory, which is referred to as the input gate.

$$p_i = sigmoid\left(W_{p,h} * h_{t-1} + W_{p,x} * X_t + b_p\right) \quad (4)$$

$$l_i = tanh\left(W_{i,h} * h_{t-1} + W_{i,x} * X_t + b_i\right) \quad (5)$$

where $p_i$ is the percentage of potential memory to remember by input gate and $l_i$ is the potential long-term memory by the input gate.

Following the acquisition of a new long-term memory, the new long-term memory transforms the tanh activation function. The resultant value then dictates the proportion of short-term memory to be propagated forward, this phase is labeled as the output gate.

$$c_t = c_{t-1} * f_l + p_i \star l_i \quad (6)$$

$$p_o = \sigma\left(W_{o,h} * h_{t-1} + W_{o,x} * x_t + b_o\right) \quad (7)$$

$$h_t = p_o * tanh\left(c_t\right) \quad (8)$$

where $c_t$ is the new long-term memory or termed as the new cell state, $p_o$ is the percentage of the potential memory to remember by output gate and $h_t$ is the new short-term memory which is the output.

LSTM networks use consistent weights and biases across stages, allowing them to handle varying data sequence lengths. At each step, LSTMs learn weight parameters $W$ and bias parameters $b$ to reduce prediction discrepancies.

*Continual Learning by EWC and LSTM (CEL)*

Initially, we segmented the disease time-series data into various distinct contexts (subsection 3.2.1). Each of these contexts represents a unique environment or domain where the aim is to predict the outbreak of a particular disease. During the training phase, we start by feeding the time series model with the first context. This model generates predictions (subsection 3.2.2, time series prediction), and the Mean Squared Error (MSE) serves as the original loss function for this first context. Mathematically, this can be represented as [40]:

$$MSE = \frac{1}{k}\sum_{m=1}^{k}(y_m - \acute{y}_m)^2 \quad (9)$$

where $y_m$ and $\acute{y}_m$ are the actual and predicted values for the $m^{th}$ instance and $k$ is the total number of instances in $C_i$.

Once the initial training on the first context is completed, the model parameters are saved, and the FIM (subsection 3.2.2, Memorizing Important Parameters) evaluates the importance of each model parameter. Essentially, FIM is calculated as the expected value of the square of the gradient of the loss function concerning each parameter. This is usually done during the backpropagation phase. The formula for a basic FIM element is represented in Eq. 1. EWC (subsection 3.2.2, EWC) calculates the penalty term. The penalty term is guided by the FIM values. Thus, for subsequent contexts, the loss function evolves into a regularized loss, as follows, that incorporates both the original loss (MSE) and a penalty term [31].

$$Regularized\ Loss = MSE + P \quad (10)$$

$$P = \sum_i \lambda/2 \left[F_i(\theta_i - \theta_{i,old}^*)^2\right] \quad (11)$$

where, $\lambda$ is a regularization parameter, and $\theta_{i,\,old}$ represents the parameters from the old contexts.

This learning process is iterative and continues for each subsequent context, and model parameters are consistently updated based on their weighted importance from previous contexts, as indicated by the FIM values. By adhering to this approach, CEL manages to adapt to new data while retaining the knowledge acquired in earlier contexts, thereby achieving a balance between adaptability and memory retention. All these steps are presented in Algorithm 1.

### 3.2.3 CEL Performance Metrics

We customized the following performance metrics [37] for the CEL evaluation. These metrics are theoretically varied and explained in our scenario.

*(i) R-squared during evaluation:* During the evaluation phase (concurrent with the training phase of sequential data), each context $C_i$ is evaluated on its respective test set after being trained on the cumulative training sets of all previous contexts $\sum_{j=1}^{i} C_j$. The R-squared for each context is calculated as:

$$R - squared_{EV,C_i} = 1 - \frac{\sum_{m=1}^{k}(y_m - \acute{y}_m)^2}{\sum_{m=1}^{k}(y_m - \bar{y})^2} \quad (12)$$

where $y_m$ and $\acute{y}_m$ are the actual and predicted values for the $m^{th}$ instance, $k$ is the total number of instances, and $\bar{y}$ denotes the mean of the observed values in the test set of $C_i$. R-squared takes any values between 0 and 1, and a higher R-squared indicates that more variability is explained by the model.

*(ii) R-squared during reevaluation:* After the model has been trained on all contexts $\sum_{i=1}^{N} C_i$ (where $N$ is the total number of contexts in the dataset), its performance is re-evaluated on the test set of each context. The R-squared during reevaluation for each context is measured similarly to the R-squared during an evaluation.

$$R - squared_{RE,C_i} = 1 - \frac{\sum_{m=1}^{k}(y_m - \acute{y}_m)^2}{\sum_{m=1}^{k}(y_m - \bar{y})^2} \quad (13)$$

where $R - squared_{RE,C_i}$ is the R-squared value of $C_i$ during reevaluation. This helps to understand if there has been catastrophic forgetting.

*(iii) R-squared Forgetting:* It measures how much performance has dropped (if at all) for each context after the model has been trained on all contexts $\sum_{i=1}^{N} C_i$. R-squared forgetting for each context $C_i$ is calculated as:

$$F_{R-squared}(C_i) = (R - squared_{EV,C_i}) - (R - squared_{RE,C_i}) \quad (14)$$

where $R - squared_{EV,C_i}$ (Eq. 12) is R-squared of $C_i$ during evaluation and $R - squared_{RE,C_i}$ (Eq. 13) is the R-squared during reevaluation.

It ranges from [-1,1], where 1 represents the higher forgetting, showing the model has forgotten everything about the $C_i$ after learning a new context, 0 implies no forgetting and -1 suggests improvement in the context $C_i$ after learning new contexts.

*(iv) Memory Stability:* Another forgetting measure is memory stability. It's the complement of the average forgetting. It indicates how stable the learning is across all contexts $\sum_{i=1}^{N} C_i$. Memory stability of R-squared can be defined as:

$$MST_{R-squared} = 1 - \frac{1}{N}\sum_{i=1}^{N} F_{R-squared}(C_i) \quad (15)$$

where $F_{R-squared}(C_i)$ is R-squared forgetting for context $C_i$ (Eq. 14). The Range of the values is [0,1], where 1 shows that the model is highly stable in remembering previous knowledge. and 0 indicates the opposite, where the model is unstable in retaining past knowledge.

---

**Algorithm 1:** The proposed CEL model.

**Input**: Context (*C*), Number of contexts (*N*), Train data ($T_r$), Test data ($T_s$), ewc_lambda (λ), Fisher information matrix (FIM).
1: LSTM model initialization
2: Set loss == MSE
3: $C_N \leftarrow$ train_test_split_contexts(*data, N*)
4: **for** $C_i$ in $C_N$ **do**
5:    $T_r, T_s \leftarrow$ load_data(*C*)
6:    **for** $C_i$ ($T_r, T_s$) **do** //evaluation
7:      **for** each feature and label in $T_r$ **do**
8:        predictions ← model(*inputs*)
9: **if** ($C_i$ == 0) **then**
10:   loss ← criterion(predictions, *labels*)
11: **else**
12:   loss ← ewc_loss(predictions, *labels*, λ, LSTM, FIM)
13: **end if, end for**
14: FIM ← compute_FIM($T_r$)
15: **for** $C_i$ ($T_r, T_s$) **do** //reevaluation
16:   **for** each feature and label in $T_s$ **do**
17:     predictions ← model(*inputs*)
18:   **end for**
19:   Compute R-squared
20: **end for**
21: **for** $C_i$ in $C_N$ **do**
22:   Compute forgetting
23: **end for**
24: Calculate memory stability
**Output**: Trained CEL model with predictions.

## 4. Experiments

### 4.1 Data Acquisition and Preprocessing

*Monkey Pox (Mpox) Dataset*

Mpox disease outbreak data was acquired from the official repository of "Our World in Data" [41]. The new daily confirmed cases from the geographical region of Africa have a timeframe from 08 May 2022 to 31 July 2023, which is approximately 15 months. Post this initial preprocessing; we derived a subset of the data termed smooth cases from refining our dataset further and making it amenable for CL model. Fig. 6(a) provides a time-series visual representation of the number of Mpox disease patients. According to the segmentation strategy, context-based data description is presented in Table 1.

*Influenza Dataset*

We sourced data on Influenza-Like Illness (ILI) rates in the United States spanning from week 30 of 2003 to week 51 of 2019 from the Centers for Disease Control and Prevention's National Center for Immunization and Respiratory Diseases via their "FluView Interactive" platform [42]. This dataset comprises approximately 832 weekly data points, representing a comprehensive temporal record over nearly 16 years. Fig. 6(b), is a time-series visual representation of these ILI rates, providing insights into overarching trends and seasonality. Table 2 presents the context-wise data description according to the segmentation strategy.

*Measles Dataset*

The dataset pertaining to measles incidence was meticulously procured from the European Centre for Disease Prevention and Control's Atlas platform [43]. This data encompasses monthly observations for the European Union (EU) region from January 2001 to December 2019, depicted in Fig. 6(c). The whole data was divided into different contexts according to the segmentation strategy; the descriptions of these contexts are given in Table 3. Prior to its integration into our continual learning model, the dataset underwent a normalization process using min-max normalization to ensure uniformity and compatibility

$$\tilde{x} = \frac{x - x_{min}}{x_{max} - x_{min}} \quad (16)$$

where $\tilde{x}$ denotes the normalized value, and $x$ represents the actual value.

### 4.2 Experimental Implementation

*Computational Resources:* The experiments were conducted on a machine equipped with an Intel(R) Core (TM) i7-6700 CPU @ 3.40GHz 3.41 GHz and 16 GB RAM. No specialized hardware like GPUs was used for the experiments.

*Parameters setting:* We employed a grid search approach to explore the hyperparameter space. The hyperparameters were selected based on their impact on the model's performance metrics, namely R-squared. The model was trained and evaluated across multiple contexts for each combination of hyperparameters. The performance was then averaged over all contexts to determine the most effective set of hyperparameters, and optimal values were found to be a learning rate of 0.01, hidden dimensions of 32, a batch size of 32, and lambda 1000.

*Software and Libraries:* The software and libraries used in the experiments are Python 3.8, PyTorch 1.9, Pandas 2.0.3, Scikit-learn 1.3.0, Matplotlib 3.7.2, Seaborn 0.12.2, NumPy 1.24.3. By adhering to this experimental setting, we ensured that the study was reproducible and the results were reliable.

**Table 1.** Contexts-based data description of Mpox data.

| Context | Start Date | End Date | Days | Mean | Std |
|---|---|---|---|---|---|
| 1 | 08-05-22 | 21-06-22 | 45 | 2.68 | 4.23 |
| 2 | 22-06-22 | 05-08-22 | 45 | 4.20 | 2.45 |
| 3 | 06-08-22 | 19-09-22 | 45 | 8.02 | 4.24 |
| 4 | 20-09-22 | 03-11-22 | 45 | 7.40 | 2.77 |
| 5 | 04-11-22 | 18-12-22 | 45 | 4.14 | 1.28 |
| 6 | 19-12-22 | 01-02-23 | 45 | 3.43 | 1.84 |
| 7 | 02-02-23 | 18-03-23 | 45 | 2.44 | 1.79 |
| 8 | 19-03-23 | 02-05-23 | 45 | 3.03 | 3.28 |
| 9 | 03-05-23 | 16-06-23 | 45 | 3.43 | 2.27 |
| 10 | 17-06-23 | 31-07-23 | 45 | 3.48 | 2.21 |

**Table 2.** Contexts-based data description of Influenza data.

| Context | Start Date | End Date | Weeks | Mean | Std |
|---|---|---|---|---|---|
| 1 | 46-2002 | 25-2004 | 84 | 0.015 | 0.014 |
| 2 | 26-2004 | May-06 | 84 | 0.015 | 0.009 |
| 3 | 22-2005 | 36-2007 | 84 | 0.014 | 0.008 |
| 4 | 37-2007 | 16-2009 | 84 | 0.018 | 0.011 |
| 5 | 17-2009 | 48-2010 | 84 | 0.021 | 0.016 |
| 6 | 49-2010 | 28-2012 | 84 | 0.016 | 0.009 |
| 7 | 29-2012 | Jul-14 | 84 | 0.019 | 0.012 |
| 8 | 32-2013 | 39-2015 | 84 | 0.017 | 0.011 |
| 9 | 40-2015 | 19-2017 | 84 | 0.020 | 0.010 |
| 10 | 20-2017 | 50-2018 | 84 | 0.020 | 0.018 |

**Table 3.** Contexts-based data description of Measles data.

| Context | Start Date | End Date | Months | Mean | Std |
|---|---|---|---|---|---|
| 1 | Aug-99 | Nov-02 | 39 | 0.080 | 0.060 |
| 2 | Dec-02 | Mar-06 | 39 | 0.238 | 0.232 |
| 3 | April-06 | Jul-09 | 39 | 0.021 | 0.010 |
| 4 | Aug-09 | Nov-12 | 39 | 0.043 | 0.042 |
| 5 | Dec-12 | Mar-16 | 39 | 0.091 | 0.070 |
| 6 | April-16 | Jul-19 | 39 | 0.395 | 0.320 |

### 4.3 Baseline Models

In our study, we compared our proposed model CEL with the following studies. We selected these studies because of their distinct CL strategies.

- *GEM:* Operating on the Gradient Episodic Memory (GEM) principle, this model combats forgetting by retaining a subset of data from earlier contexts. Its model uses MLP layers and ensures that model updates don't adversely affect performance on previously stored data by suitably constraining gradient updates [26].
- *XdG:* This model uses a gating mechanism within its RNN structure by utilizing the Context-dependent Gating (XdG) strategy. It modulates RNN activations based on the context at hand. It tailors its learning to the current context without compromising knowledge from previous contexts by "gating" specific activations [32].

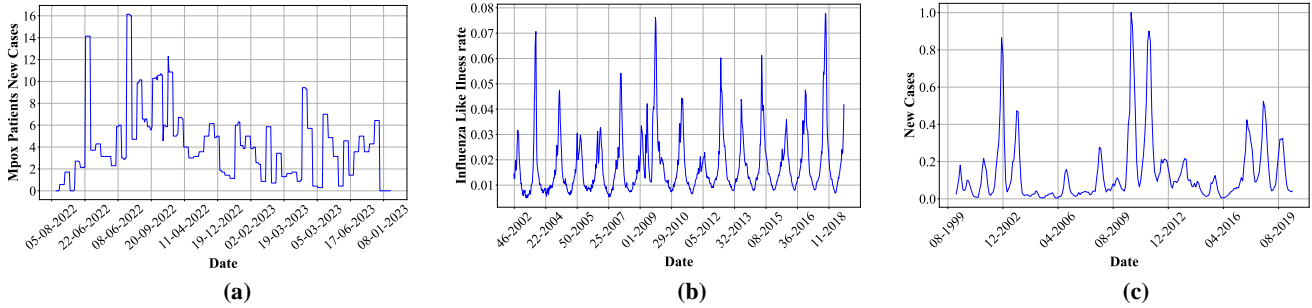

**Figure 6.** Visualization of different disease data, **(a)** daily Mpox cases in Africa, **(b)** weekly Influenza-like illness rate in the US, **(c)** monthly Measles cases in the EU.

- *EMC:* EMC uses episodic memory systems and prediction-error-driven memory consolidation. Its model architecture uses adaptive architecture and utilizes the LSTM layers to mitigate catastrophic forgetting [25].

All baseline models were evaluated against CEL under consistent experimental settings.

## 5. Results and Discussion

In disease outbreak prediction, the ability of a model to adapt to new data without forgetting previously learned patterns is paramount. This work evaluated the proposed model CEL and compared it with GEM [26], XdG [32], and EMC [25].

### 5.1 Outbreak Predictions and Comparisons

For the Mpox and Influenza $N = 10$ and Measles $N = 6$ contexts were selected by segmentation strategy, then for these disease outbreak predictions, all models, GEM [26], XdG [32], EMC [25], and CEL were sequentially trained across each context. Subsequent to training on each training set of contexts, each model was assessed on its corresponding test set utilizing Eq. 12. Reevaluation was performed following the completion of the training across all contexts based on Eq. 13. Additionally, forgetting ($F_{R-squared}$) using Eq. 14 and memory stability based on Eq. 15 were also calculated. A comparative analysis of the performance metrics of these models is presented in subsequent sections.

#### 5.1.1 Mpox Prediction

During the evaluation phase, the R-squared values were computed and depicted in Fig. 7(a), indicating the models' aptitude in learning the training data. A noticeable enhancement in performance was observed across all models during this phase; CEL consistently delivered outstanding R-squared scores across multiple contexts. Other models, also demonstrating commendable R-squared values, were occasionally outperformed by CEL. Upon reevaluation, the R-squared values were again calculated and presented in Fig. 7(b). It was observed that, apart from CEL, all other models demonstrated diminished performance. For instance, during task 7, CEL achieved the highest reevaluation score of 0.3454, whereas GEM recorded the lowest score of -0.2192. This suggests that while these models were proficient during the training phase, they appeared to forget information from prior contexts when subjected to training in subsequent contexts. Further analysis was conducted to assess the models' retention capabilities illustrated in Fig. 7(c), wherein CEL was found to exhibit the least amount of forgetting, as a comparative analysis revealed that CEL outperformed EMC by 69% at task 3, underscoring its superior retention abilities.

#### 5.1.2 Influenza Prediction

CEL displayed elevated R-squared values across every context during evaluation, suggesting its adeptness in adapting to the training Influenza data, as demonstrated in Fig. 7(d). Interestingly, EMC and XdG manifested progressive adaptability, with their efficacy becoming increasingly apparent in later contexts. When these models were reevaluated, CEL maintained its proficiency, underscoring its robust generalization capabilities, presented in Fig. 7(e). For example, during task 3, XdG obtained the lowest score at 0.5005, whereas CEL exhibited a notable peak at 0.8344. Despite their strong training performance during evaluation, other models exhibited a drop during reevaluation, proving their potential overfitting during training. The forgetting analysis Fig. 7(f) revealed CEL's prowess in retaining knowledge from previous contexts, with minimal deterioration in performance. In contrast, all other models displayed more pronounced forgetting in contexts, suggesting high forgetting in maintaining the previous knowledge. For example, in task 3, CEL exhibited an 88% higher proficiency in preserving information from prior contexts compared to the average performance of all other models.

#### 5.1.3 Measles Prediction

CEL demonstrated a diverse performance, with notably high R-squared values during the training phase, when models were trained on the Measles dataset, as illustrated in Fig. 7 (g). Other models also showed consistent and commendable performance across all contexts. These models were reevaluated to evaluate in terms of retention and adaptability. Results presented in Fig. 7(h) show that CEL's performance indicated its robustness in generalizing. The robustness of this performance is evident in task 3, where CEL (0.6245) outperformed all other models by 481%, showcasing a substantially superior performance compared to the average score of -0.1636, which indicates an inconsistent outcome. These results exhibited a decline in showing some degree of forgetting in all other models except CEL, suggesting potential challenges in retaining knowledge. The forgetting analysis in Fig. 7(i) further highlights the performance of the models in terms of forgetting. It shows that CEL was relatively stable throughout the context, while other models showed most forgetting. For example, in task 5, CEL exhibited proficiency (0.0716) in memorizing previous contexts, whereas XdG displayed a context-forgetting rate of 0.4892.

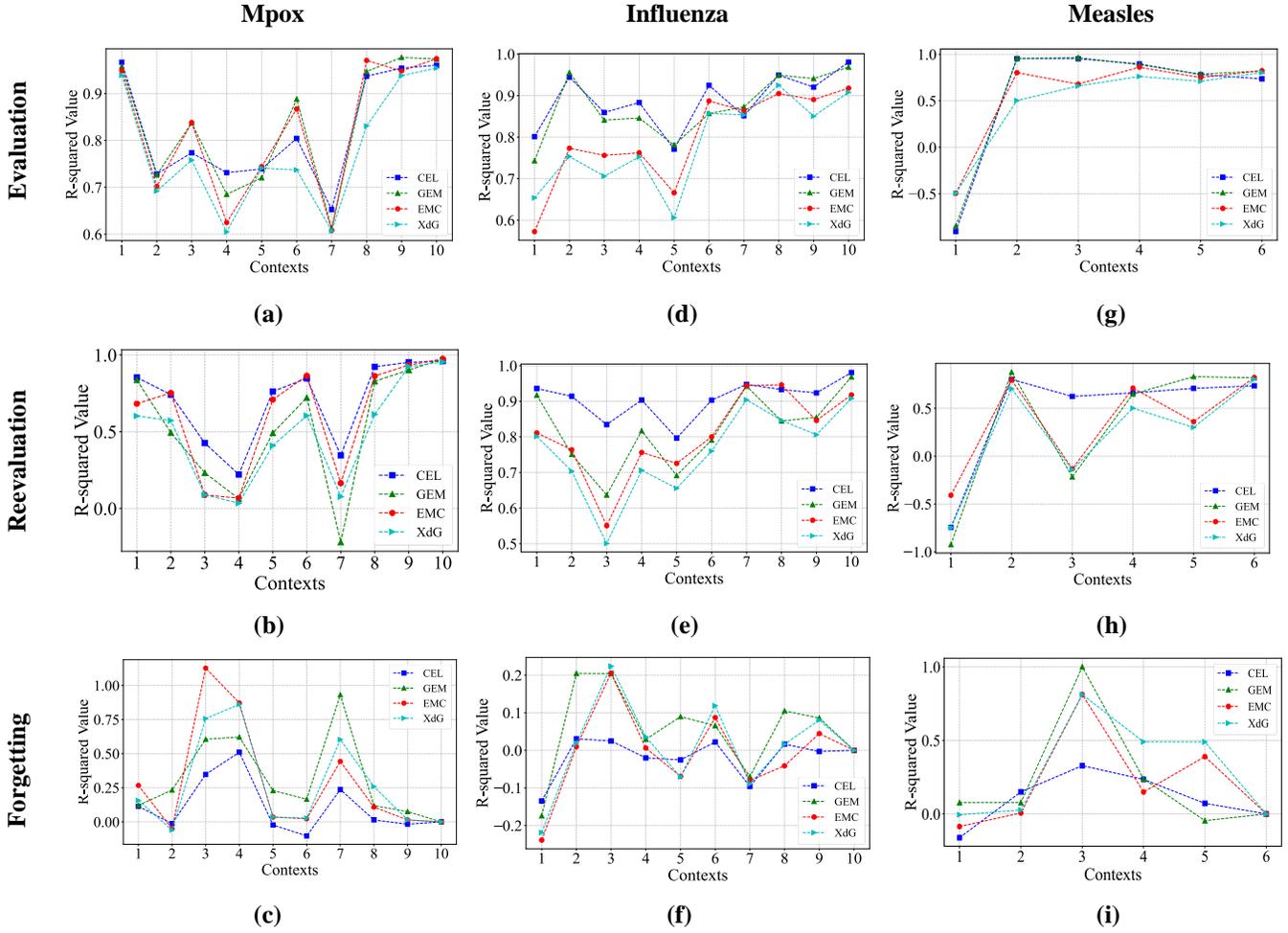

**Figure 7.** R-squared trends during Evaluation, Reevaluation, and Forgetting measure **(a)** evaluation of Mpox, **(b)** reevaluation of Mpox, **(c)** forgetting of Mpox, **(d)** evaluation of Influenza, **(e)** reevaluation of Influenza, **(f)** forgetting of Influenza, **(g)** evaluation of Measles, **(h)** reevaluation of Measles, **(i)** forgetting of Measles.

Across all three diseases, the proposed model, CEL, consistently demonstrated robust and stable performance, both during the evaluation and reevaluation phases. GEM, while generally performing well, faced challenges in specific contexts, indicating potential areas for optimization. EMC and XdG, though showing promise in several contexts, had their struggles, especially in the initial contexts for each disease.

In analyzing the average trend of forgetting across all tasks within three disease datasets for four models, it is evident that the proposed CEL model demonstrates the 65% lowest average forgetting, comparatively to other models.

Regarding memory stability, a crucial metric in continual learning, CEL emerged with superior performance, as in Table 4. Our findings indicate that in the Mpox data, CEL demonstrated a 39% superior resilience in retaining knowledge compared to XdG. Furthermore, it proved to be 17% and 19% more effective in ensuring memory stability than the average performance of other models in the Influenza and Measles datasets, respectively. The analysis of overall memory stability becomes evident through a comparison of the average performances of each model across all disease datasets, with the following hierarchical order: $CEL > GEM > EMC > XdG$.

**Table 4.** Comparisons of memory stability of different models.

| Disease | GEM [26] | EMC [25] | XdG [32] | CEL |
|---|---|---|---|---|
| Mpox | 0.70615 | 0.68987 | 0.61876 | **0.86213** |
| Influenza | 0.90875 | 0.89720 | 0.80613 | **0.96265** |
| Measles | 0.75972 | 0.73042 | 0.66217 | **0.84259** |

### 5.2 FIM Impact on LSTM Parameters

In CEL's disease outbreak prediction model, the Fisher Information Matrix (FIM) assumes a pivotal role, offering insights into how distinct model components contribute to the prediction of disease outbreaks. We thoroughly examine FIM's significance, indicating its impact on various facets of the model and its broader implications in addressing the challenges of Domain-IL.

**Table 5.** Summary of LSTM parameters.

| Parameter Name | No. of Parameters | Transformation Type |
|---|---|---|
| lstm.weight_ih_l0 | 1536 | Input to Hidden |
| lstm.weight_hh_l0 | 4096 | Hidden to Hidden |
| lstm.bias_ih_l0 | 128 | Input to Hidden |
| lstm.bias_hh_l0 | 128 | Hidden to Hidden |
| linear.weight | 32 | Fully Connected |
| linear.bias | 1 | Fully Connected |

In our LSTM-based model, several critical elements drive its functionality. These encompass parameters are presented in Table 5. These parameters, characterized by their unique shapes, undergo iterative adjustments during training to minimize prediction errors, and the FIM serves as a guiding compass, unveiling which elements exhibit heightened responsiveness to changes when confronted with different disease datasets. Those with higher FIM values are better equipped to assimilate new information; thus, the model allocates maximum focus to them.

Similarly, Elastic Weight Consolidation (EWC) plays a crucial role in preventing the model from erasing previously acquired knowledge as it learns new information. FIM plays a critical role in determining the protection level required for each component through EWC. It introduces a regularization term into the base loss, resulting in a regularized loss, as shown in Figs. 8-10 for all diseases. This regularized loss is initially higher at the beginning of context training but gradually decreases with the introduction of new contexts. So, the model's performance becomes stable after the last context. Iterations also show high trends where standard deviation abruptly changes for contexts. For example, in the Mpox for contexts 3 and 5. Components deemed more critical due to their higher FIM values receive the highest protection, preserving vital knowledge while permitting other facets of the model to evolve.

To visually represent the evolution of different model components, we employ 3D graphs in Figs. 11-13. These visualizations clearly show which components remain stable and which undergo transformation when faced with new disease data. The 3D plots map Fisher information across several contexts and parameter indices, providing insights into the shifting significance of LSTM parameters across varying contexts.

As the model encounters an increasing number of contexts, the divergence in parameter influence becomes more distinct. Certain parameters rise in importance, signifying their adaptability across diverse contexts. In contrast, some parameters, notably among bias terms like *lstm.bias_ih_l0* and *lstm.bias_hh_l0*, appear to diminish in influence, hinting at potential redundancies. This observation underscores the notion that not all parameters hold equal significance across all contexts.

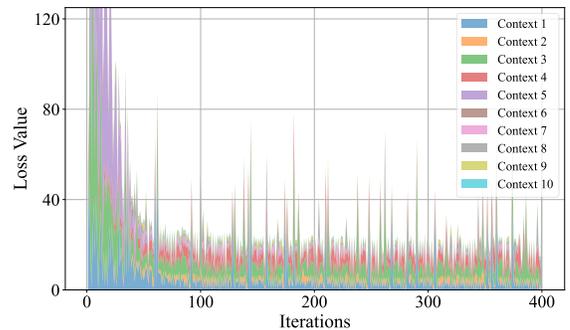

**Figure 8.** Regularized loss over different Mpox contexts by proposed CEL model.

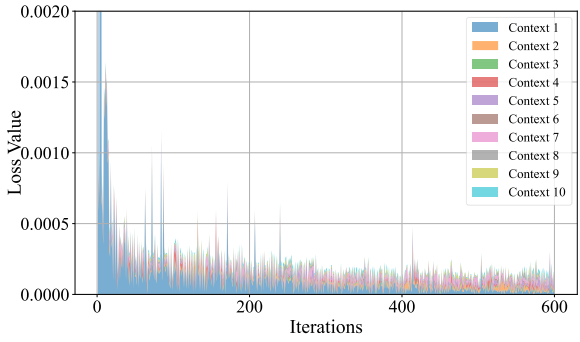

**Figure 9.** Regularized loss over different Influenza contexts by proposed CEL model.

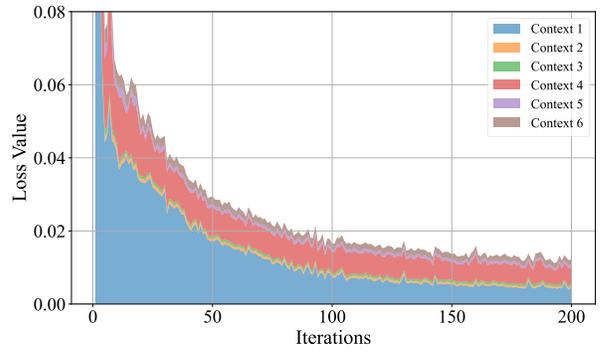

**Figure 10.** Regularized loss over different Measles contexts by proposed CEL model.

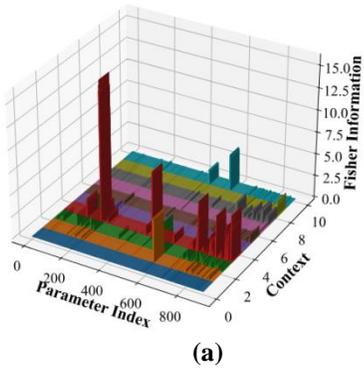

(**a**)

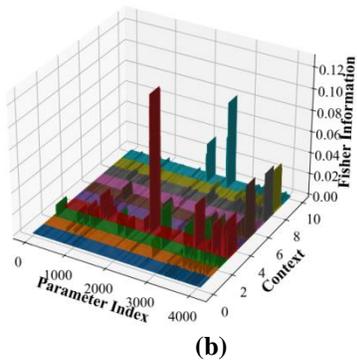

(**b**)

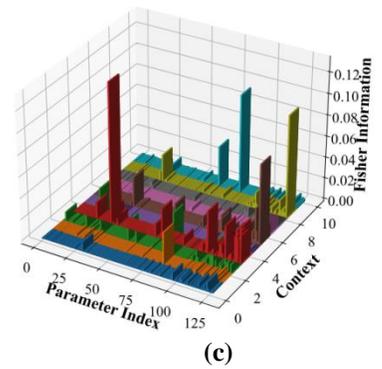

(**c**)

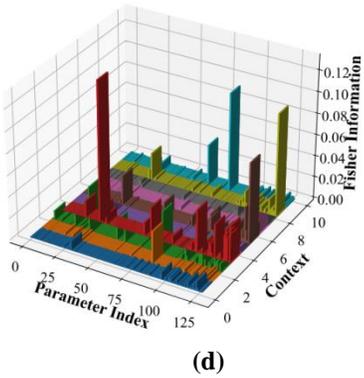

(**d**)

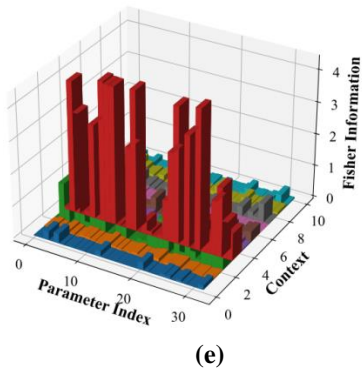

(**e**)

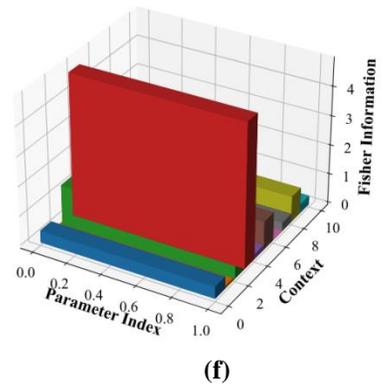

(**f**)

**Figure 11.** Fisher Information Changes for different parameters during the training of Mpox data by CEL, (**a**) lstm.weight_ih_10, (**b**) lstm.weight_hh_10, (**c**) lstm.bias_ih_10, (**d**) lstm.bias_hh_10, (**e**) linear.weight, and (**f**) linear.bias.

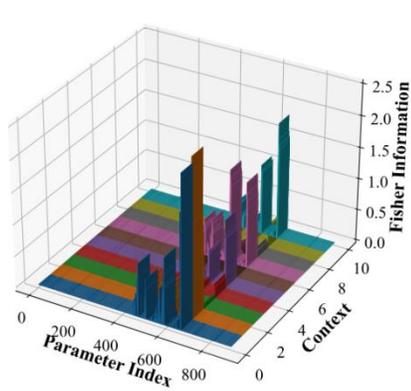 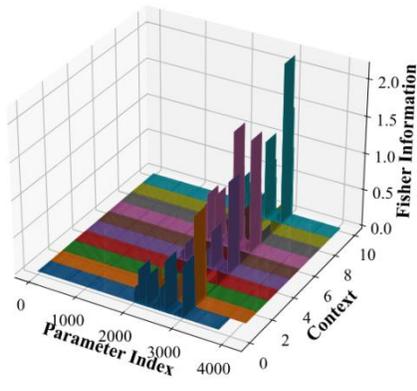 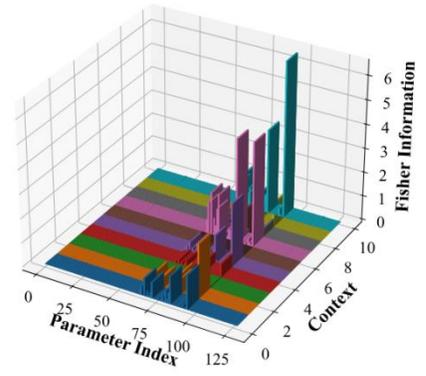

(a) (b) (c)

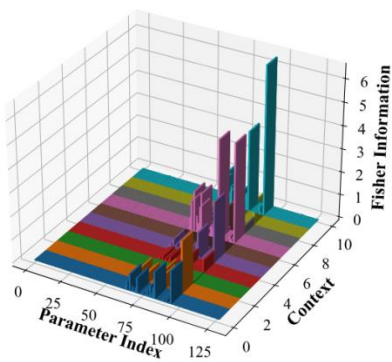 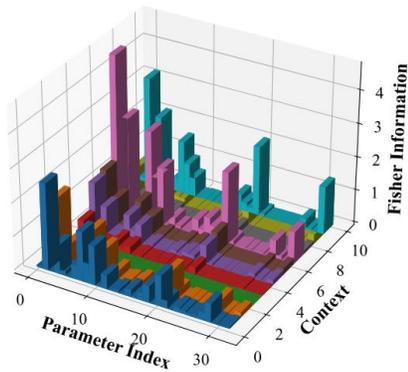 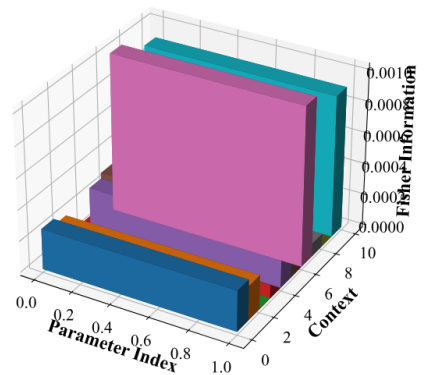

(d) (e) (f)

**Figure 12**. Fisher Information Changes for different parameters during the training of Influenza data by CEL, **(a)** lstm.weight_ih_l0, **(b)** lstm.weight_hh_l0, **(c)** lstm.bias_ih_l0, **(d)** lstm.bias_hh_l0, **(e)** linear.weight, and **(f)** linear.bias.

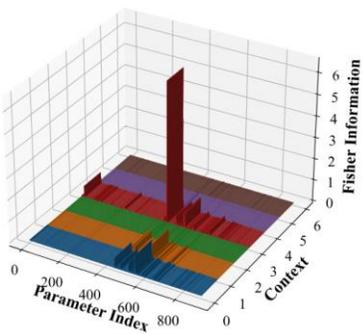 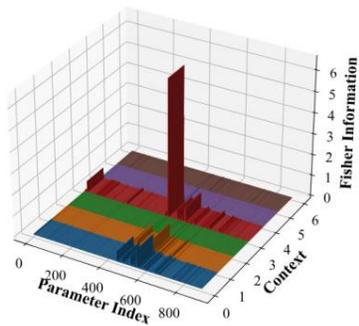 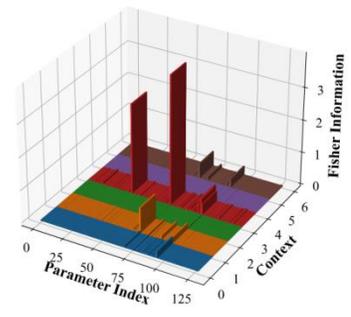

(a) (b) (c)

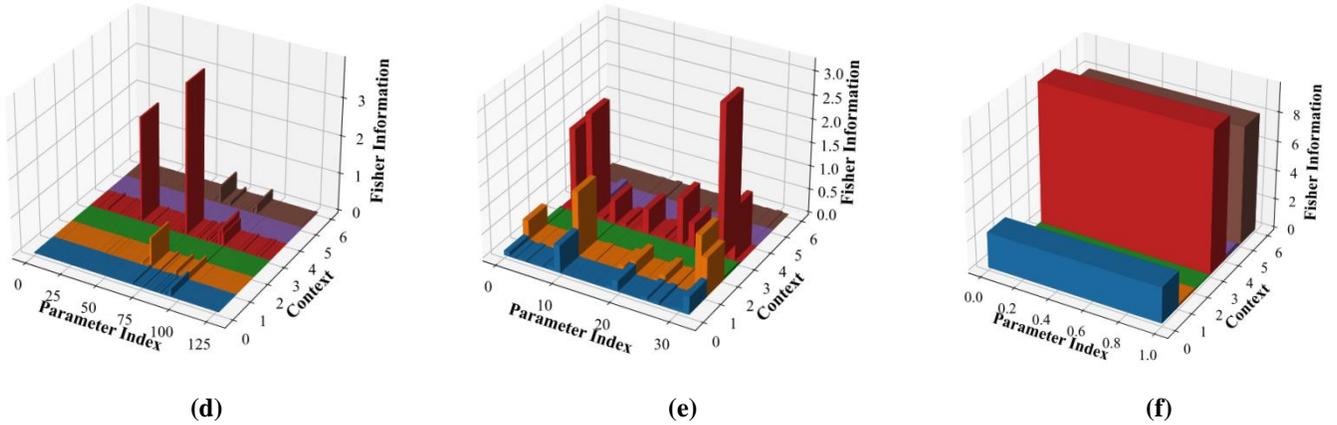

| | | | |
|---|---|---|---|
| (d) | (e) | (f) |

**Figure 13.** Fisher Information Changes for different parameters during the training of Measles data by CEL, **(a)** lstm.weight_ih_10, **(b)** lstm.weight_hh_10, **(c)** lstm.bias_ih_10, **(d)** lstm.bias_hh_10, **(e)** linear.weight, and **(f)** linear.bias.

### 5.3 Segmentation Strategy Impact

One significant factor guiding our segmentation strategy is the average R-squared values, which offer a valuable metric for understanding the impact of different context divisions on model performance. We opted for a finer segmentation strategy for the Mpox (Mx) and Influenza (I) outbreaks, which exhibited relatively higher average R-squared values across the 10th context, presented with bold values in Table 6. This choice was based on the observation that more contexts allowed our models to capture the nuances in the data distribution, resulting in improved predictive performance. Conversely, for Measles (M), which displayed a much lower R-squared with a higher number of contexts (with minimum on context 7 (0.01) to max 0.577 on 10, which is even lower than 60%), a coarser segmentation into 6th contexts (highlighted with bold values) was preferred as it is giving 0.657 value. This segmentation choice ensured that we did not unnecessarily burden the models with an excessive number of contexts while still enabling effective learning.

Determining the optimal number of contexts (N) relies on some principles. Observing changes in memory requirements with varying *N* strikes a balance between computational efficiency and model accuracy. In addition, leveraging more historical data is crucial for the effective capturing of underlying patterns. Furthermore, this ensures that the segmentation strategy aligns with data characteristics and practical computational considerations, contributing to an optimized strategy for enhanced model performance across diverse outbreaks.

**Table 6.** Proposed CEL's performances across different numbers of contexts for disease (D); Mpox (Mx), Influenza (I), Measles (M) and comparison with GEM [26], XdG [32], and EMC [25].

| Models | D | Number of Contexts | | | | |
|---|---|---|---|---|---|---|
| | | 6 | 7 | 8 | 9 | 10 |
| GEM | Mx | 0.722 | 0.698 | 0.355 | 0.712 | **0.805** |
| XdG | | 0.601 | 0.58 | 0.31 | 0.689 | **0.751** |
| EMC | | 0.634 | 0.72 | 0.325 | 0.702 | **0.78** |
| CEL | | 0.728 | 0.729 | 0.358 | 0.718 | **0.818** |
| GEM | I | 0.844 | 0.817 | 0.809 | 0.861 | **0.887** |
| XdG | | 0.711 | 0.71 | 0.78 | 0.7 | **0.8** |
| EMC | | 0.775 | 0.742 | 0.812 | 0.786 | **0.776** |
| CEL | | 0.847 | 0.817 | 0.827 | 0.867 | **0.887** |
| GEM | M | **0.555** | 0.022 | 0.444 | 0.477 | 0.475 |
| XdG | | **0.451** | 0.021 | 0.413 | 0.388 | 0.307 |
| EMC | | **0.473** | 0.01 | 0.423 | 0.405 | 0.336 |
| CEL | | **0.657** | 0.029 | 0.457 | 0.498 | 0.577 |

## 6. Conclusion

In this study, we introduced and evaluated the CEL model, a novel approach to continual learning that synergistically combines Elastic Weight Consolidation (EWC) with LSTM networks. The primary motivation behind CEL is to address the catastrophic forgetting phenomenon of DNNs, a significant challenge in continual learning, especially in the domain of disease outbreak prediction.

Our extensive experiments across three distinct diseases–Mpox, Influenza, and Measles, provided valuable insights into the efficacy of the CEL model. The results consistently demonstrated that CEL offers a robust and stable performance across varying contexts, outpacing other state-of-the-art models like (GEM, XdG, and EMC). Notably, the CEL model's ability to maintain high R-squared values in evaluation and reevaluation coupled with 65% minimal R-squared forgetting and 18% higher memory stability (Table 4), highlights its potential as a reliable tool for disease outbreak prediction. The consistent performance of CEL across different diseases highlights its versatility and adaptability. By leveraging the memory retention capabilities of EWC and the sequential data processing strengths of LSTM, the CEL model offers a balanced approach that captures both the temporal patterns and the evolving nature of disease data.

The CEL model, while promising, exhibits certain technical weaknesses and areas for future improvement. Firstly, despite the valuable insights provided by R-squared utilized in assessing the regression model, exclusive reliance on this metric is cautioned against. The inclusion of RMSE is recommended. Secondly, CEL's effectiveness may vary depending on the characteristics of disease data, particularly when diseases demonstrate irregular or unpredictable patterns. Lastly, the model might face challenges when confronted with abrupt and significantly different context changes, such as emerging diseases with distinct patterns.

**Data Availability:** The data for this work is available at https://github.com/aI-area/CEL.

**Acknowledgments:** This work is supported by the National Natural Science Foundation of China (62273322), the Strategic Priority CAS Project (XDB38050100), and the Shenzhen Science and Technology Program (KQTD20190929172835662).

**Conflicts of Interest:** The authors declare no conflict of interest.

**Author Contributions:** Saba Aslam: Data curation, Methodology, Software, Writing - original draft, Writing - review & editing, Visualization. Abdur Rasool: Writing - review & editing, Visualization, Formal analysis, Validation. Hongyan Wu: Conceptualization, Funding, Methodology, Supervision, Resources, Formal analysis, Project administration. Xiaoli Li: Formal analysis, Visualization. All authors have read and agreed to the published version of the manuscript.

## References

[1] R. Singh and R. Singh, "Applications of sentiment analysis and machine learning techniques in disease outbreak prediction – A review," *Materials Today: Proceedings,* vol. 81, pp. 1006-1011, 2023/01/01/ 2023, doi: https://doi.org/10.1016/j.matpr.2021.04.356.

[2] J. Zhang, P. Zhou, Y. Zheng, and H. Wu, "Predicting influenza with pandemic-awareness via Dynamic Virtual Graph Significance Networks," *Computers in Biology and Medicine,* vol. 158, p. 106807, 2023.

[3] D. Feng and H. Chen, "A small samples training framework for deep Learning-based automatic information extraction: Case study of construction accident news reports analysis," *Advanced Engineering Informatics,* vol. 47, p. 101256, 2021.

[4] R. Hadsell, D. Rao, A. A. Rusu, and R. Pascanu, "Embracing change: Continual learning in deep neural networks," *Trends in cognitive sciences,* vol. 24, no. 12, pp. 1028-1040, 2020.

[5] D. Kudithipudi *et al.*, "Biological underpinnings for lifelong learning machines," *Nature Machine Intelligence,* vol. 4, no. 3, pp. 196-210, 2022.

[6] Z. Chen and B. Liu, *Lifelong machine learning*. Springer, 2018.

[7] G. I. Parisi, R. Kemker, J. L. Part, C. Kanan, and S. Wermter, "Continual lifelong learning with neural networks: A review," (in eng), *Neural Netw,* vol. 113, pp. 54-71, May 2019, doi: 10.1016/j.neunet.2019.01.012.

[8] C. S. Lee and A. Y. Lee, "Clinical applications of continual learning machine learning," *The Lancet Digital Health,* vol. 2, no. 6, pp. e279-e281, 2020.

[9] M. Perkonigg *et al.*, "Dynamic memory to alleviate catastrophic forgetting in continual learning with medical imaging," *Nature Communications,* vol. 12, no. 1, p. 5678, 2021/09/28 2021, doi: 10.1038/s41467-021-25858-z.

[10] E. Verwimp *et al.*, "CLAD: A realistic Continual Learning benchmark for Autonomous Driving," *Neural Networks,* vol. 161, pp. 659-669, 2023/04/01/ 2023, doi: https://doi.org/10.1016/j.neunet.2023.02.001.

[11] K. Doshi and Y. Yilmaz, "Continual learning for anomaly detection in surveillance videos," in *Proceedings of the IEEE/CVF conference on computer vision and pattern recognition workshops*, 2020, pp. 254-255.

[12] D. Rolnick, A. Ahuja, J. Schwarz, T. Lillicrap, and G. Wayne, "Experience replay for continual learning," *Advances in Neural Information Processing Systems,* vol. 32, 2019.

[13] J. Ramapuram, M. Gregorova, and A. Kalousis, "Lifelong generative modeling," *Neurocomputing,* vol. 404, pp. 381-400, 2020.

[14] R. Aljundi, M. Lin, B. Goujaud, and Y. Bengio, "Gradient based sample selection for online continual learning," *Advances in neural information processing systems,* vol. 32, 2019.

[15] J. Zhang *et al.*, "Class-incremental learning via deep model consolidation," in *Proceedings of the IEEE/CVF Winter Conference on Applications of Computer Vision*, 2020, pp. 1131-1140.

[16] G. M. van de Ven, T. Tuytelaars, and A. S. Tolias, "Three types of incremental learning," *Nature Machine Intelligence,* vol. 4, no. 12, pp. 1185-1197, 2022.

[17] D. Lopez-Paz and M. A. Ranzato, "Gradient episodic memory for continual learning," *Advances in neural information processing systems,* vol. 30, 2017.

[18] M. De Lange *et al.*, "A continual learning survey: Defying forgetting in classification tasks," *IEEE transactions on pattern analysis and machine intelligence,* vol. 44, no. 7, pp. 3366-3385, 2021.

[19] Z. Mai, R. Li, J. Jeong, D. Quispe, H. Kim, and S. Sanner, "Online continual learning in image classification: An empirical survey," *Neurocomputing,* vol. 469, pp. 28-51, 2022.

[20] M. Masana, X. Liu, B. Twardowski, M. Menta, A. D. Bagdanov, and J. Van De Weijer, "Class-incremental learning: survey and performance evaluation on image classification," *IEEE Transactions on Pattern Analysis and Machine Intelligence,* vol. 45, no. 5, pp. 5513-5533, 2022.

[21] M. J. Mirza, M. Masana, H. Possegger, and H. Bischof, "An efficient domain-incremental learning approach to drive in all weather conditions," in *Proceedings of the IEEE/CVF Conference on Computer Vision and Pattern Recognition*, 2022, pp. 3001-3011.

[22] A. Kara, "Multi-step influenza outbreak forecasting using deep LSTM network and genetic algorithm," *Expert Systems with Applications,* vol. 180, p. 115153, 2021/10/15/ 2021, doi: https://doi.org/10.1016/j.eswa.2021.115153.

[23] S. Yang and Y. Bao, "Comprehensive learning particle swarm optimization enabled modeling framework for multi-step-ahead influenza prediction," *Applied Soft Computing,* vol. 113, p. 107994, 2021.

[24] N. Wu, B. Green, X. Ben, and S. O'Banion, "Deep transformer models for time series forecasting: The influenza prevalence case," *arXiv preprint arXiv:2001.08317,* 2020.

[25] G. Schillaci, U. Schmidt, and L. Miranda, "Prediction Error-Driven Memory Consolidation for Continual Learning: On the Case of Adaptive Greenhouse Models," *KI-Künstliche Intelligenz,* vol. 35, pp. 71-80, 2021.

[26] S. K. Amalapuram, A. Tadwai, R. Vinta, S. S. Channappayya, and B. R. Tamma, "Continual learning for anomaly based network intrusion detection," in *2022 14th International Conference on COMmunication Systems & NETworkS (COMSNETS)*, 2022: IEEE, pp. 497-505.

[27] S.-A. Rebuffi, A. Kolesnikov, G. Sperl, and C. H. Lampert, "icarl: Incremental classifier and representation learning," in *Proceedings of the IEEE conference on Computer Vision and Pattern Recognition*, 2017, pp. 2001-2010.

[28] H. Shin, J. K. Lee, J. Kim, and J. Kim, "Continual learning with deep generative replay," *Advances in neural information processing systems,* vol. 30, 2017.

[29] Z. Li and D. Hoiem, "Learning without forgetting," *IEEE transactions on pattern analysis and machine intelligence,* vol. 40, no. 12, pp. 2935-2947, 2017.

[30] R. Aljundi, F. Babiloni, M. Elhoseiny, M. Rohrbach, and T. Tuytelaars, "Memory aware synapses: Learning what (not) to forget," in *Proceedings of the European conference on computer vision (ECCV)*, 2018, pp. 139-154.

[31] J. Kirkpatrick *et al.*, "Overcoming catastrophic forgetting in neural networks," *Proceedings of the national academy of sciences,* vol. 114, no. 13, pp. 3521-3526, 2017.

[32] N. Y. Masse, G. D. Grant, and D. J. Freedman, "Alleviating catastrophic forgetting using context-dependent gating and synaptic stabilization," *Proceedings of the National Academy of Sciences,* vol. 115, no. 44, pp. E10467-E10475, 2018.

[33] J. Rajasegaran, M. Hayat, S. H. Khan, F. S. Khan, and L. Shao, "Random path selection for continual learning," *Advances in Neural Information Processing Systems,* vol. 32, 2019.

[34] J. Xu and Z. Zhu, "Reinforced continual learning," *Advances in Neural Information Processing Systems,* vol. 31, 2018.

[35] A. Mallya and S. Lazebnik, "Packnet: Adding multiple tasks to a single network by iterative pruning," in *Proceedings of the IEEE conference on Computer Vision and Pattern Recognition*, 2018, pp. 7765-7773.

[36] M. Masana, T. Tuytelaars, and J. Van de Weijer, "Ternary feature masks: zero-forgetting for task-incremental learning," in *Proceedings of the IEEE/CVF conference on computer vision and pattern recognition*, 2021, pp. 3570-3579.

[37] A. Chaudhry, P. K. Dokania, T. Ajanthan, and P. H. Torr, "Riemannian walk for incremental

learning: Understanding forgetting and intransigence," in *Proceedings of the European conference on computer vision (ECCV)*, 2018, pp. 532-547.

[38] A. Aich, "Elastic weight consolidation (EWC): Nuts and bolts," *arXiv preprint arXiv:2105.04093,* 2021.

[39] R. A. Fisher, "On the mathematical foundations of theoretical statistics," *Philosophical transactions of the Royal Society of London. Series A, containing papers of a mathematical or physical character,* vol. 222, no. 594-604, pp. 309-368, 1922.

[40] Z. Wang and A. C. Bovik, "Mean squared error: Love it or leave it? A new look at signal fidelity measures," *IEEE signal processing magazine,* vol. 26, no. 1, pp. 98-117, 2009.

[41] "Edouard Mathieu, Fiona Spooner, Saloni Dattani, Hannah Ritchie and Max Roser (2022) - "Mpox (monkeypox)". Published online at OurWorldInData.org. Retrieved from: 'https://ourworldindata.org/monkeypox' [Online Resource]."

[42] "Centers for Disease Control and Prevention, National Center for Immunization and Respiratory Diseases (NCIRD), FluView interactive,.." https://www.cdc.gov/flu/weekly/fluviewinteractive.htm.

[43] "European Centre for Disease Prevention and Control's Atlas platform." https://atlas.ecdc.europa.eu/public/index.aspx?Dataset=27&HealthTopic=20.